%% file: main.tex
\documentclass[conference]{IEEEtran}
\IEEEoverridecommandlockouts
\usepackage{cite}
\usepackage{amsmath,amssymb,amsfonts}
\usepackage{algorithmic}
\usepackage{graphicx}
\usepackage{textcomp}
\usepackage{xcolor}
\usepackage{caption}
\usepackage{placeins}
\usepackage{url} 
\usepackage{bbm}
\usepackage{booktabs} 
\usepackage{xcolor}
\usepackage{amssymb}
\usepackage{multirow}
\def\BibTeX{{\rm B\kern-.05em{\sc i\kern-.025em b}\kern-.08em
    T\kern-.1667em\lower.7ex\hbox{E}\kern-.125emX}}

\begin{document}

\title{CAT-3DGS Pro: A New Benchmark for Efficient 3DGS Compression}

\author{
Yu-Ting Zhan\textsuperscript{1}, He-bi Yang\textsuperscript{1}, Cheng-Yuan Ho\textsuperscript{1}, Jui-Chiu Chiang\textsuperscript{2}, and Wen-Hsiao Peng\textsuperscript{1}\\ \textsuperscript{1}National Yang Ming Chiao Tung University, Taiwan\\ \textsuperscript{2}National Chung Cheng University, Taiwan
}

\maketitle

\input{Sections/0_Abstract}
\input{Sections/1_Introduction}

% \input{Sections/2_RelatedWork}
\input{Sections/3_Method}

\input{Sections/4_Experiment}

\input{Sections/5_Conclusion}

\input{Sections/Reference}
\end{document}

%% file: Sections/0_Abstract.tex
\begin{abstract}

3D Gaussian Splatting (3DGS) has shown immense potential for novel view synthesis. However, achieving rate-distortion-optimized compression of 3DGS representations for transmission and/or storage applications remains a challenge. CAT-3DGS introduces a context-adaptive triplane hyperprior for end-to-end optimized compression, delivering state-of-the-art coding performance. Despite this, it requires prolonged training and decoding time. To address these limitations, we propose CAT-3DGS Pro, an enhanced version of CAT-3DGS that improves both compression performance and computational efficiency. First, we introduce a PCA-guided vector-matrix hyperprior, which replaces the triplane-based hyperprior to reduce redundant parameters. To achieve a more balanced rate-distortion trade-off and faster encoding, we propose an alternate optimization strategy (A-RDO). Additionally, we refine the sampling rate optimization method in CAT-3DGS, leading to significant improvements in rate-distortion performance. These enhancements result in a 46.6\% BD-rate reduction and 3× speedup in training time on BungeeNeRF, while achieving 5× acceleration in decoding speed for the \textit{Amsterdam} scene compared to CAT-3DGS.
\end{abstract}
\vspace{1mm}

%% file: Sections/1_Introduction.tex
\section{Introduction}
3D Gaussian Splatting (3DGS)~\cite{b1} has emerged as a promising 3D scene representation, excelling in novel view synthesis within differentiable rendering frameworks due to its real-time, high-quality rendering. Despite its strengths, the redundancy inherent in 3DGS has spurred research efforts to develop more compact representations~\cite{b2,b3,b4,b5,b6,b7,b8,b9,b10,b11,b12,b13}. However, the efficient transmission of compressed 3DGS remains an overlooked challenge, requiring entropy coding and rate-distortion optimization. 

Recently, the rate-distortion-optimized compression for 3DGS ~\cite{b14,b15,b16,b17,b18,b19,b20} started to gain attention. Its aim is to balance the bit rate and rendering quality in an end-to-end manner. Specifically, HAC~\cite{b15}, ContextGS~\cite{b17}, HEMGS~\cite{b19}, and CAT-3DGS~\cite{b18} utilize the ScaffoldGS~\cite{b7} representation, adopting an anchor-based approach where each anchor point encodes a group of Gaussian primitives into a latent feature vector and signals their geometry offsets relative to the anchor position, as well as a 6-dimensional scaling factor. The latent feature vector is further quantized by a scalar quantizer for entropy coding. These early works leverage the hyperprior idea from learned image compression~\cite{b21} to model their coding probabilities. In forming the hyperprior, HAC~\cite{b15} constructs multi-scale binary hash grids. ContextGS learns a quantized feature for each anchor. HEMGS adopts the pre-trained PointNet++ and an instance-aware hash grid to generate the hyperprior. Notably, to utilize spatial redundancy for better coding efficiency, both ContextGS and HEMGS additionally resort to spatial autoregressive contextual coding. ContextGS hierarchically organizes anchor points into three successive coding levels, while HEMGS compresses anchor points in raster order by referring to their decoded neighbors. Rather than explicitly organizing anchor points, CAT-3DGS projects them onto multi-scale triplanes as the hyperprior, and encodes these triplanes using spatial autoregressive models. Although achieving state-of-the-art coding results, CAT-3DGS exhibits long training and decoding time. 
\input{Sections/1-1_teaser}

%CAT-3DGS~\cite{b25}, inspired by tensor decomposition~\cite{b8}, employs content-adaptive triplanes to construct the hyperprior. 

This paper introduces CAT-3DGS Pro, an enhanced version of CAT-3DGS that addresses its long training time and high decoding complexity while boosting its coding performance through an improved training strategy. The key advancements include the adoption of a PCA-guided vector-matrix (VM) hyperprior, which reduces the computational complexity of CAT-3DGS while retaining its capability to represent the 3D space effectively. This VM hyperprior significantly accelerates both training and decoding processes. To further increase parallelism, we additionally divide a high-resolution hyperprior plane into 4 smaller ones for separate entropy encoding/decoding. We also implement an alternative optimization technique to strike a better balance between the rate-distortion performance and encoding speed. These improvements culminate in state-of-the-art compression performance, as shown in Figure\ref{fig:teaser}. Remarkably, CAT-3DGS Pro achieves a 5× speedup in decoding speed on the \textit{Amsterdam} scene and a 3× speedup in training time on BungeeNeRF compared to CAT-3DGS.

%refined training strategies, as detailed in Sec.\ref{sec:rdo_training}, including  

%% file: Sections/1-1_teaser.tex
\begin{figure}[t]
    \centering
    \resizebox{\linewidth}{!}{
    \includegraphics[width=0.49\textwidth]{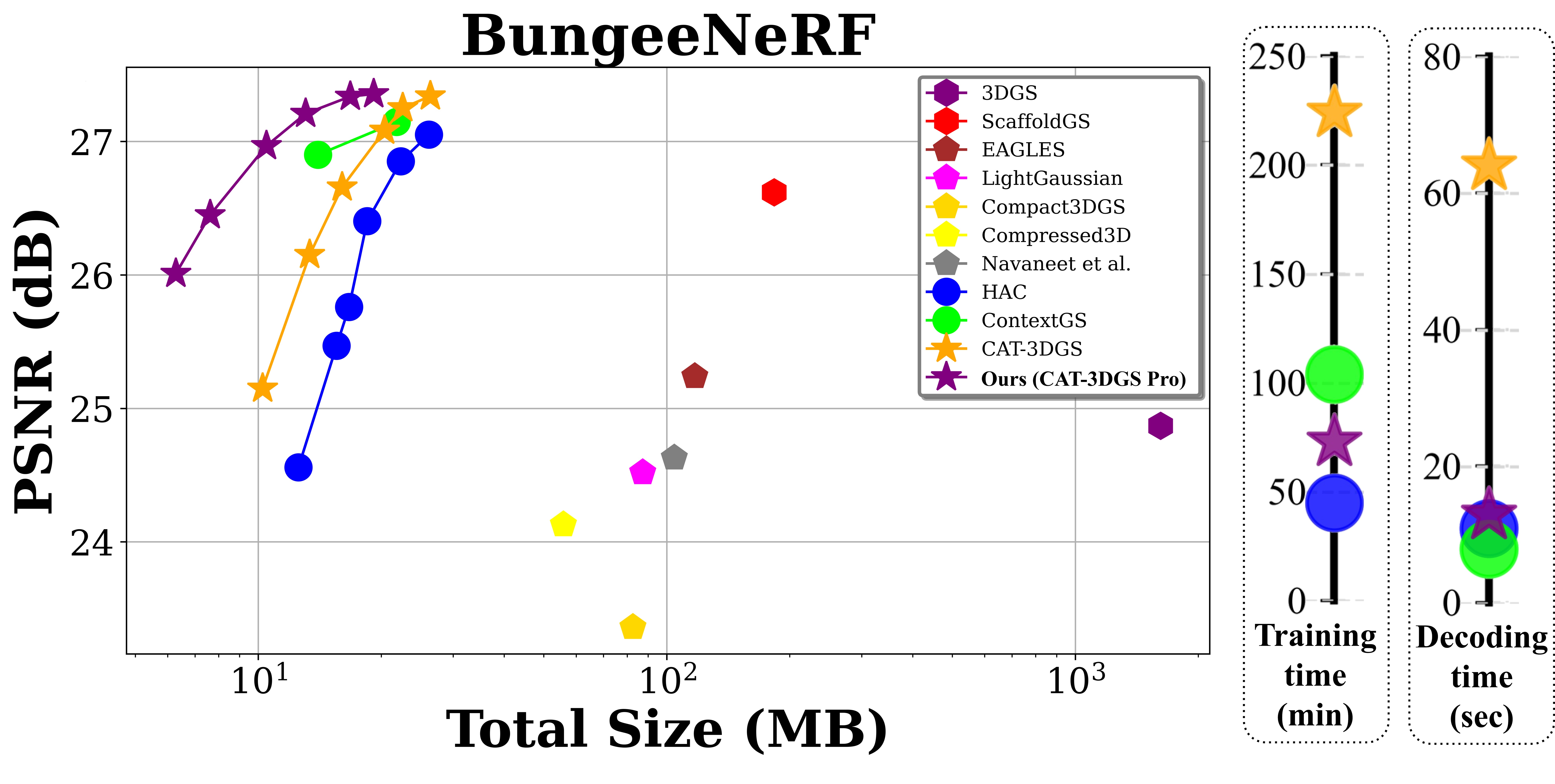}  
    }
    \caption{Comparison of rate-distortion performance, training time on the BungeeNeRF dataset, and decoding time for the \textit{Amsterdam} scene.}
    \label{fig:teaser}
\end{figure}

%% file: Sections/3_Method.tex
\input{Sections/3-1_overview}
\section{Proposed Method: CAT-3DGS Pro}
\label{sec:motivation}
%\vspace{-5.0mm}
Figure~\ref{fig:overview} depicts the architecture of our CAT-3DGS Pro. The encoding process for a 3D scene begins with constructing a Scaffold-GS representation based on anchor points. These anchor points are characterized by their positions $\boldsymbol{x} \in \mathbb{R}^3$ and attributes $\mathcal{A}$, which include the latent features $\boldsymbol{f} \in \mathbb{R}^{50}$, offsets $\{\boldsymbol{O_i} \in \mathbb{R}^{3}\}_{i=1}^K$, and scaling factors $\boldsymbol{l} \in \mathbb{R}^6$ (part (a)). To entropy encode/decode the attributes of the anchors, a PCA-guided VM hyperprior (parts (c)(d)) is queried using the PCA-transformed coordinates $\boldsymbol{x'}=(x',y',z')$ of an anchor point to retrieve $g(x',y'),\gamma(z')$ for decoding the probability distributions of its attributes (part (e)). For encoding/decoding the latent features $\boldsymbol{f}$, we additionally leverage a channel-wise contextual coding scheme \cite{b25} to exploit the \textit{intra correlation} among their components (part (f)). The VM hyperprior consists of multi-scale planes $\mathcal{P}_{x'y'}$ aligned with the two principal components bearing the largest eigenvalues, alongside a positionally encoded vector in the direction of the remaining principal component. These multi-scale planes $\mathcal{P}_{x'y'}$ also require encoding/decoding. They are quantized and encoded using a spatial autoregressive model (part (g)). In summary, the compressed bitstream for CAT-3DGS Pro includes $\mathcal{P}_{x'y'}$, the anchors' attributes $\mathcal{A}$ and positions $\boldsymbol{x}$, the binary mask, and the network weights. The binary mask implements a view frequency-aware masking mechanism \cite{b18} to skip Gaussian primitives with minimal impact on rendering quality. The anchors' positions $\boldsymbol{x}$ and network weights are represented in 16-bit and 32-bit floating-point formats, respectively, while the binary mask undergoes entropy coding. The rendering with Gaussian splatting simply follows Scaffold-GS~\cite{b7}.
%Following Scaffold-GS, we set $K=10$. Additionally, a view frequency-aware masking mechanism \cite{b25} is integrated to selectively mask Gaussian primitives with minimal impact on rendering quality. 

%To build the learnable, vector-matrix (VM) hyperprior for modeling the probability distributions of anchor attributes, we conduct a principal component analysis of the anchor points' positions $\boldsymbol{x}$ (Figure \ref{fig:overview} (c)). Along the two dimensions with the highest eigenvalues, a number of multi-scale planes (i.e. matrices) $\mathcal{P}_{xy}$ are created to form part of the hyperprior (Figure \ref{fig:overview} (d)). The remaining dimension perpendicular to the multi-scale plans is a vector, along which positional encoding is applied to complete the hyperprior construction (Figure \ref{fig:overview} (d)). The vector and multi-scale planes/matrices jointly form the Vector-Matrix (VM) hyperprior, which is queried using the PCA-transformed coordinates $(x',y',z')$ of an anchor point to retrieve $g(x',y'),\gamma(z')$ for decoding the probability distributions.  

%The $\mathcal{P}_{xy}$ planes are quantized and encoded using a spatial autoregressive model (Sec.~\ref{sec:plane_AR}). . 

% An MLP decoder $F_S$ decodes the latent feature of an anchor $\boldsymbol{\hat{f}}$ to retrieve color, opacity, rotation, and scale of the associated Gaussian primitives. The decoded offsets $\{\boldsymbol{\hat{O}_i}\}_{i=1}^K$ and scaling $\boldsymbol{\hat{l}}$ are combined with the anchor position $\boldsymbol{x}$ to reconstruct their positions.

\subsection{PCA-Guided Vector-Matrix (VM) Hyperprior}
\label{sec:hyperprior}

For entropy encoding/decoding anchor attributes, we employ a PCA-guided VM hyperprior to predict their probability distributions. In Figure \ref{fig:overview} (c), PC1, PC2, and PC3 are the three principal components extracted from anchors' locations $\boldsymbol{x}$, where the majority of anchors are predominantly distributed along PC1 and PC2, which have the highest eigenvalues. With CAT-3DGS Pro, we reduce the multi-scale triplane structure of CAT-3DGS to multi-scale planes $\mathcal{P}_{x'y'}$ oriented along PC1 and PC2. We hypothesize that the attributes of most anchors can already be predicted well by querying and decoding the corresponding latent features $g(x',y')$ derived from these multi-scale planes (i.e. dense-grid representations). These planes of various scales ensure that both the coarse and fine details are well captured. To distinguish between the $x'y'$-planes of different scales, we augment $\boldsymbol{\mathcal{P}}_{x'y'}$ with an upsampling scale $r$, denoted as $\boldsymbol{\mathcal{P}}_{r,x'y'} \in \mathbb{R}^{ch \times rB \times rB}$, where $ch$ denotes the number of channels, $r$ represents the upsampling scales, and $B$ indicates the spatial resolution of the $x'y'$ planes at the base scale (i.e., $r=1$). To address the information not captured by the multi-scale planes, we further introduce a positionally encoded vector $r(z')$ that takes the $z'$ coordinate along PC3 as input. The attributes $\mathcal{A}$ of anchor points are treated as a vector-valued function defined in 3D space. Conceptually, our hyperprior VM approximates this function by a vector-matrix decomposition. 

%The PCA presents a transformation of the anchor's position $\boldsymbol{x}$ into $\boldsymbol{x'}$. This is followed by a contraction function~\cite{b3} to map potentially unbounded positions into bounded ones: 
%\begin{equation} \boldsymbol{x'} = \text{contract}(\frac{\boldsymbol{R^x} (\boldsymbol{x} - \boldsymbol{\mu^x})}{\boldsymbol{\sigma^x}}), 
%\end{equation} where $\boldsymbol{\mu^x} \in \mathbb{R}^3$ represents the mean vector, $\boldsymbol{\sigma^x} \in \mathbb{R}^3$ are the variances along the three principal axes, and $\boldsymbol{R^x} \in \mathbb{R}^{3 \times 3}$ is the PCA rotation matrix. 

%To formulate this design, we pre-process the 3D space by performing principal component analysis (PCA) on the anchor positions $\boldsymbol{x}$, followed by a contraction function~\cite{b3} to map potentially unbounded positions onto bounded planes
%After pre-processing, the positions $\boldsymbol{x'}$, where $\boldsymbol{x'} = (x', y', z')$, are such that the $x'$ component captures the greatest variance, indicating the primary orientation, while $z'$ corresponds to the dimension with the least variance.

%To elucidate the design of our vector-matrix-based hyperprior (Figure~\ref{fig:overview} (d)), we begin by selecting the two principal components with the highest variance to construct the $xy$ planes, denoted as $\boldsymbol{\mathcal{P}}_{xy}$. These components are instrumental in capturing the most significant spatial features from the data, allowing us to efficiently encapsulate the essential aspects of the 3D space while simultaneously reducing complexity. 

To retrieve the plane feature $g(x',y')$ for an anchor point at $\boldsymbol{x'}$, the coordinates $(x', y')$ are first projected onto each 2D plane $\boldsymbol{\mathcal{P}}_{r,x'y'}$. The resulting projected 2D coordinates are denoted by $\pi_{r}(x',y')$. If $\pi_{r}(x',y')$ produces fractional coordinates, interpolation is performed between the nearest integer grid points using an interpolation kernel $\psi$. In symbols, we have $\psi(\boldsymbol{\mathcal{P}} _{r, x'y'}, \pi_{r}(x',y'))$. The interpolated features obtained from all planes are then concatenated to yield $g(x',y')$:
\begin{equation}
%\begin{split}
g(x',y')=\bigcup_r \psi(\boldsymbol{\mathcal{P}} _{r, x'y'}, \pi_{r}(x',y')).
%\end{split}
\label{eq:hyperprior}
\end{equation}

Likewise, to get $\gamma(z')$,  we transform $z'$ into a sequence of sinsodial Fourier features $\gamma(z')$ across $L=10$ frequencies: 
\begin{equation}
% \scriptsize
\small
\gamma(z') = \left( \sin(2^0 \pi z', \cos(2^0 \pi z'), \ldots, \sin(2^{L-1} \pi z'), \cos(2^{L-1} \pi z') \right).
\label{eq:fourier_features}
\end{equation}
Unlike the costly dense-grid representation for $\boldsymbol{\mathcal{P}}_{x'y'}$, which can effectively represent high-frequency signals, this Fourier feature vector is designed to represent smoother signal variations along the $z'$ axis.  

\label{sec:entropy_coding}
Finally, to entropy encode (or decode) the quantized attributes $\boldsymbol{\hat{a}} \in \{ \boldsymbol{\hat{f}_1}, \{\boldsymbol{\hat{O}_i}\}, \boldsymbol{\hat{l}}\}$), we concatenate the plane feature $g(x', y')$ and Fourier features $\gamma(z')$. These concatenated features are then fed into the MLP decoder $F_h$ to predict their means, variances, and quantization step size. That is, $(\boldsymbol{\mu}, \boldsymbol{\sigma}, q) = F_h([g(x', y'), \gamma(z')])$ (Figure ~\ref{fig:overview} (e)), where [·] denotes the concatenation operation. Notably, each of these attributes is assumed to follow a Gaussian distribution, with their coding probabilities given by
%\vspace{-3.2mm}
\begin{equation}
    \label{eq:hyperprior_f1}
    p(\boldsymbol{\hat{a}}| g(x', y'), \gamma(z')))=\int_{\boldsymbol{\hat{{a}}}-\frac{q}{2}} ^{\boldsymbol{\hat{{a}}}+\frac{q}{2}} \mathcal{N}(\boldsymbol{a}; \boldsymbol{\mu}, \boldsymbol{\sigma}) \, d\boldsymbol{a}. 
    %\text{with }  (\boldsymbol{\mu}, \boldsymbol{\sigma}, q) = F_{tri}(h(\boldsymbol{x'})).
\end{equation} 
%\vspace{-1.4mm}

\vspace{-2.4mm}
\subsection{Spatial Autoregressive Models (SARM) for Plane Coding}
\label{sec:plane_AR}
Given that $\boldsymbol{\mathcal{P}}_{r, x'y'}$ functions as part of the hyperprior, and are designed to capture the \textit{inter-correlations} among the attributes of anchor points in 3D space, we apply a spatial autoregressive model $F_{ARM}$ to encode the multi-scale planes in $\boldsymbol{\mathcal{P}}_{r, x'y'}$. The model $F_{ARM}$ is shared across scales $r$ and channels $ch$.
For entropy encoding (and decoding) a grid point $y_{i, j, k}$ in $\boldsymbol{\mathcal{P}}_{r, x'y'}$ within channel $k$, a context is constructed from the decoded grid points in its neighborhood, defined as $c_{i, j, k}=[\hat{y}_{i-1, j-1:j+1, k}; \hat{y}_{i, j-1, k}]$ . Using this context, $F_{ARM}$ predicts the Laplace parameters $(\mu_{i, j, k}, \sigma_{i, j, k})$ that model the distribution of $y_{i, j, k}$ (Figure~\ref{fig:overview} (g)).

% The coding probability of the quantized grid point $\hat{y}_{i, j, k}$ is then given by
% \vspace{-1.5mm}
% \begin{equation}
% \small
%     %\begin{split}
%     \label{eq:spatialarm}
%     p(\hat{y}_{i, j, k}|c_{i, j, k})= \int_{\hat{y}_{i, j, k}-\frac{Q}{2}}^{\hat{y}_{i, j, k}+\frac{Q}{2}} Laplace(\mu_{i, j, k}, \sigma_{i, j, k}) \, d y_{i, j, k},
% \end{equation}
% % \vspace{-5.5mm}
% where $Q=1/16$ is the quantization step size.  

\vspace{-1mm}

\subsection{Rate-distortion Optimization}
\label{sec:rdo_training}
The training of CAT-3DGS Pro involves minimizing the rate-distortion cost ${L}_\text{Scaffold} + \lambda_r {L}_\text{rate}$ together with a masking loss $\lambda_m {L}_m$: 
\begin{equation}
    \label{eq:loss}
    {L} = {L}_\text{Scaffold} + \lambda_r {L}_\text{rate} + \lambda_m {L}_m,
\end{equation}
where we follow Scaffold-GS~\cite{b7} to evaluate ${L}_\text{Scaffold}$, which includes the distortion between the original and rendered images as well as a regularization term applied to the scales $s$ of Gaussian primitives. $L_{m}$ is the mask loss adopted from CAT-3DGS~\cite{b18} to regularize the view frequency-aware masking.
${L}_\text{rate}$ indicates the number of bits needed to signal the hyperprior and the anchors' attributes: \begin{equation}
    \begin{split}
     \label{eq:rate_loss}
        L_\text{rate} = \frac{1}{N(50 + 6 + 3K)}(L_{\text{rate}}^{\boldsymbol{\mathcal{A}}} + \lambda_\text{tri} L_{\text{rate}}^{\boldsymbol{\mathcal{P}}}),
    \end{split}
\end{equation}
where $L_{\text{rate}}^{\boldsymbol{\mathcal{A}}} = - \sum_{\boldsymbol{\hat{a}}} log_2\,p(\hat{\boldsymbol{a}})$ is the estimated bit rate of the anchors' attributes, $L_{\text{rate}}^{\boldsymbol{\mathcal{P}}} =  - \sum_{\hat{y}}log_2\,p(\hat{y})$ is the hyperprior's bit rate, and $N(50 + 6 + 3K)$ is the total number of the parameters of anchors' attributes. Following Scaffold-GS, we set $K=10$.

% In \cite{b30}, the training is the process for over-fitting the network to a video sequence, in which the samples of each steps are from the same sequence, and the code, i.e., the INR model parameters, is the same set of parameters for all steps. Thus, it is not necessary to update the rate term in every step, especially when a significant amount of computation or memory is needed for this due to the use of entropy models. an efficient training process is used, where rate $R$ and distortion $D$ are optimized alternately $D$ is minimized for the first $N$ steps, and $R$ is minimized at the $N+1$-th step.
\input{Sections/3-2_observation}

\input{Sections/4-1_RDresults}

Recognizing that the estimation of the bit rate in SARM significantly prolongs the training time, we propose an alternate rate-distortion optimization strategy, termed A-RDO. Inspired by NVRC\cite{b22}, A-RDO evaluates the rate term $L_{\text{rate}}^{\boldsymbol{\mathcal{P}}}$ periodically with an interval of $T=4$ steps, while the other loss terms are evaluated at every step. Our A-RDO significantly reduces training time while achieving better rate-distortion performance (Secs.~\ref{exp:training_time} and \ref{tab:boc_analysis}).

Additionally, we observe that the training strategy in CAT-3DGS, known as Visible Sampling Rate Optimization (ViSRO), randomly samples 5\% of the anchor points that are visible in the training views to estimate the bit rate of the anchor attributes $L_{\text{rate}}^{\boldsymbol{\mathcal{A}}}$. This approach results in a higher bit rate for representing the attributes of anchor points located in less frequently observed (peripheral) regions, which significantly limits the rate-distortion performance (Figure \ref{fig:vis_analysis} (a)). It is crucial to note that the compressed bitstream must encode all the anchor points, whether they are located in peripheral or central regions. To tackle this problem, we have revised the training strategy to evaluate the rate term independently of an anchor point's visibility.The bit rate estimation is no longer restricted to visible anchor points. This new training strategy is called Uniform Sampling Rate Optimization (USRO). Figure \ref{fig:vis_analysis} compares the bit-rate distributions of ViSRO and USRO. The results demonstrate that with USRO, central anchor points utilize a relatively higher bit rate than peripheral ones, leading to significantly improved rate-distortion performance (Sec.~\ref{exp:training_strategy}).

%% file: Sections/3-1_overview.tex
\begin{figure*}[t]
    \centering
    \centerline{\includegraphics[width=1\textwidth]{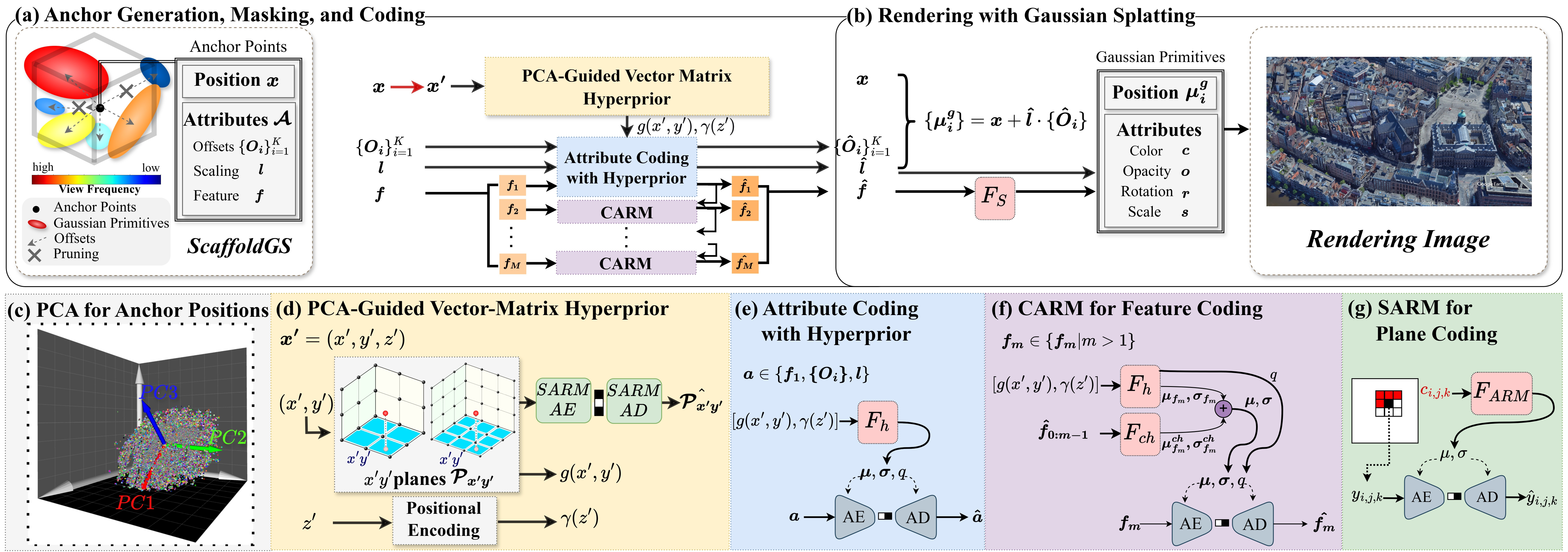}}
    % \vspace{-4mm}
    \caption{System overview of our CAT-3DGS Pro. CARM: Channel-wise Autoregressive Models. SARM: Spatial Autoregressive Models.}
    \label{fig:overview}
    % \vspace{-1.em}
\end{figure*}
% \vspace{-1.4cm}

%% file: Sections/3-2_observation.tex
\begin{figure}[t]
    \centering
    \resizebox{\linewidth}{!}{
    \includegraphics[width=0.5\textwidth]{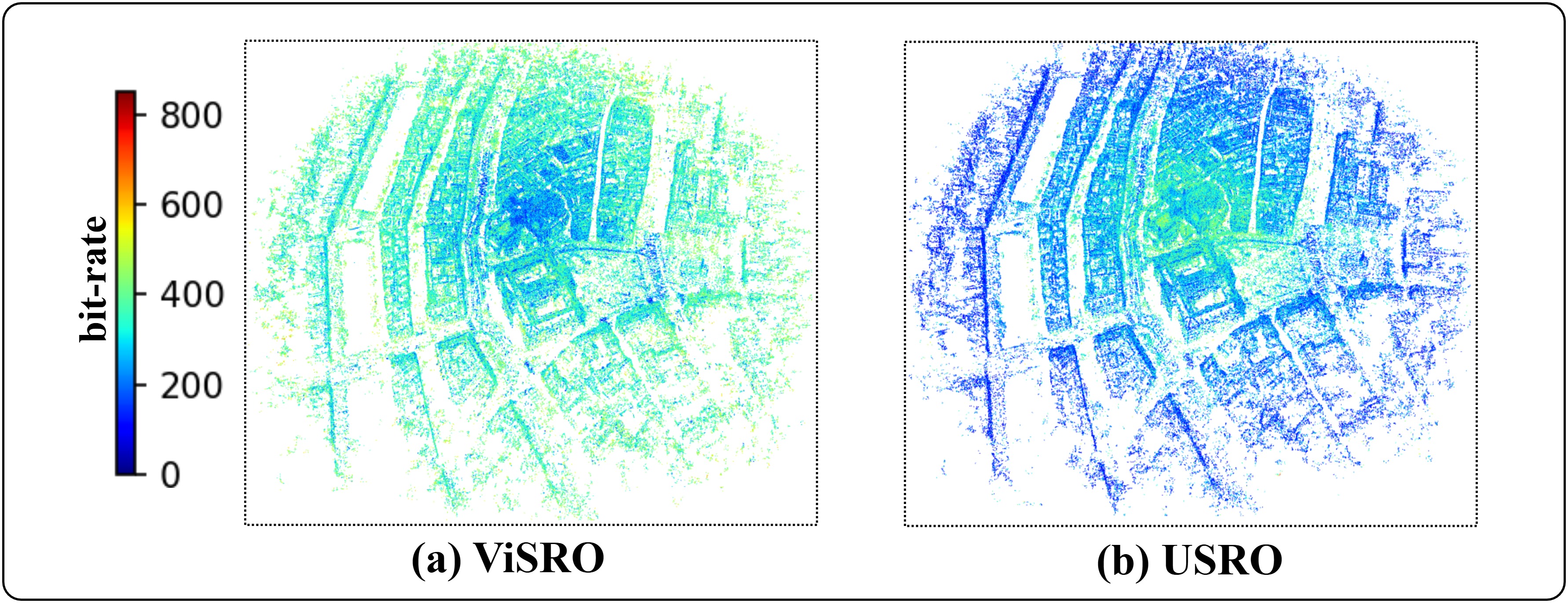}  
    }
    \caption{Visualization of the bit-rate distribution on the \textit{Amsterdam} scene for ViSRO and USRO. The bit rate indicates the number of bits needed to represent the anchor attributes of each anchor point.}
    \label{fig:vis_analysis}
\end{figure}

%% file: Sections/4-1_RDresults.tex
\begin{figure*}[ht]
    \centering
    \centerline{\includegraphics[width=0.99\linewidth]{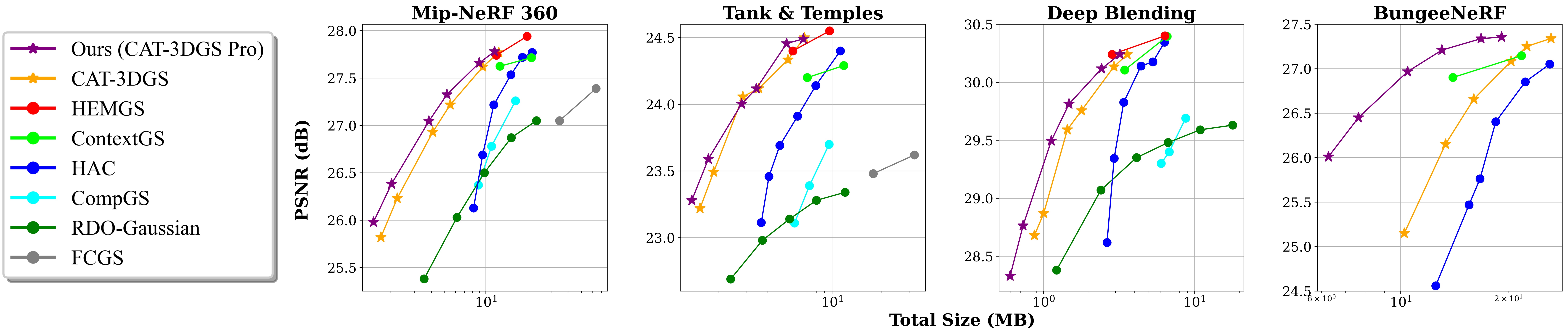}}
    
    \caption{Rate-distortion comparison of our CAT-3DGS Pro and several baseline methods.}
    \label{fig:rdresults}
\end{figure*}

%% file: Sections/4_Experiment.tex
\vspace{-0.7mm}
\section{Experimental Results}
\subsection{Implementation Details}

This section summarizes the implementation details. First, the spatial resolution $B$ of the base-scale plane ($r=1$) is determined proportionally based on the number of anchor points obtained after 10k training iterations, with $B=128$ for the highest anchor count in all training scenes and $B=64$ for the lowest one. Our multi-scale planes consists of only two scales, $r=1,2$, each with $ch=8,$ as opposed to 72 used in CAT-3DGS. %For the third principal component’s positional encoding, we set the number of levels $L$ to 10.
To accelerate the decoding process, the high-resolution plane $\boldsymbol{\mathcal{P}}_{2,x'y'}$ is split into four non-overlapping, low-resolution sub-planes, each having a spatial resolution the same as $\boldsymbol{\mathcal{P}}_{1,x'y'}$. This setup results in a total of five planes that are decoded in parallel. For rate-distortion optimization, the rate parameter $\lambda_r$ ranges from 0.002 to 0.04, and from 0.002 to 0.03 for BungeeNeRF, with $\lambda_{tri}$ fixed at 10 and $\lambda_m = \text{max}(10^{-3}, 0.3\cdot\lambda_r)$. %For A-RDO, we set $T=4$ for updating the rate of planes.

% The choices of the other hyperparameters include: $\epsilon=0.01$ ($0.0004$ for BungeeNeRF) for the view frequency-aware masking, $m = 4$ with uneven slices $(5, 10, 15, 25)$ for the channel-wise autoregressive coding.

\subsection{Rate-Distortion Comparison}
To demonstrate the rate-distortion performance of CAT-3DGS Pro, we compare it against several rate-distortion-optimized 3DGS compression schemes~\cite{b15,b14,b18,b16,b17,b19,b20}. Following the common test protocol, we evaluate their coding performance on real-world scenes, including Mip-NeRF 360~\cite{b23}, Tanks \& Temples~\cite{b24}, Deep Blending~\cite{b25}, and BungeeNeRF~\cite{b26}. As shown in Figure~\ref{fig:rdresults}, our CAT-3DGS Pro achieves the state-of-the-art rate-distortion performance, demonstrating significant improvements over CAT-3DGS primarily due to the USRO training strategy (Sec.~\ref{exp:training_strategy}). While HEMGS performs comparably to our method at high rates, it requires the computationally expensive pre-trained PointNet++ for hyperprior modeling. 

%and a raster-order spatial context autoregressive model, though no complexity metrics were reported.

\subsection{Complexity Comparison}
\input{Sections/4-2_training_time_table}

\input{Sections/4-3_decoding}

\subsubsection{Training Time}
\label{exp:training_time}
Table~\ref{table:training_time} compares the training time of several rate-distortion-optimized methods using BungeeNeRF, which is a particularly challenging large-scale dataset. Remarkably, CAT-3DGS Pro achieves a 3.1× speedup over CAT-3DGS. This improvement comes from two key advancements. First,  A-RDO reduces the training time of CAT-3DGS (with the triplane hyperprior) by 25.6\%, due to a reduced frequency of bit rate evaluations for SARM. Second, replacing the triplane hyperprior with our VM hyperprior further cuts training time by 55.8\%, primarily due to lower encoding complexity and a significant reduction in hyperprior parameters from 17M to 0.6M. 

\subsubsection{Decoding Time and Rendering Throughput}
Table~\ref{table:decode_time} compares the decoding time and rendering throughput for two scenes: \textit{Amsterdam} in BungeeNeRF, which has a high anchor count, and \textit{room} in Mip-NeRF360, which features a low anchor count. In terms of decoding time, CAT-3DGS Pro achieves approximately a 5× speedup over CAT-3DGS on the \textit{Amsterdam} scene, and demonstrates comparable decoding speed to HAC and ContextGS. However, CAT-3DGS Pro achieves much better rate-distortion performance than HAC and ContextGS. Additionally, CAT-3DGS Pro has dramatically reduced the decoding time for the multi-scale planes ($\boldsymbol{\mathcal{P}}$) from 47.4s to 2.2s. This improvement is attributed to the VM-based hyperprior representation and the parallel decoding of the multi-scale planes. Furthermore, the VM-based hyperprior features a smaller memory footprint, allowing more anchor points to be decoded in a batch. This accounts for the decoding speedup in the attribute part ($\boldsymbol{\mathcal{A}}$). Lastly, CAT-3DGS Pro achieves a higher rendering throughput than CAT-3DGS, as fewer anchor points are required to maintain similar rendering quality, primarily due to the USRO scheme. It tends to regularize the rate loss term by masking high-bit-rate anchor points while ensuring a minimal impact on distortion.
%To further analyze the contributions of each modification to this acceleration, we break down the improvements as follows: In CAT-3DGS, plane decoding is performed through sequential entropy encoding, processing the largest plane first, followed by the smallest. To accelerate plane decoding, we implement a spatial parallel scheme, which partitions the largest plane into four smaller planes that can be decoded in parallel. This optimization reduces plane ($\boldsymbol{\mathcal{P}}$) decoding time by 21\%. Furthermore, replacing the triplane hyperprior in CAT-3DGS with the PCA-guided vector-matrix results in a 94.0\% reduction in plane decoding time. Lastly, the simplified hyperprior design enables an increase in batch size, which defines the number of anchor points decoded in parallel. This modification increases the batch size from 500 to 1000 without incurring out-of-memory issues. This reduces anchor attribute ($\boldsymbol{\mathcal{A}}$) decoding timeby 38.4\%.
 
\subsection{Ablation Study}
 
\subsubsection{Impact of Hyperprior and A-RDO on Compression Performance}
\label{exp:boc_analysis}
\input{Sections/4-4_ablation_bitstream_table}
We investigate the impact of A-RDO and the VM hyperprior on compression performance, using the \textit{Amsterdam} scene in BungeeNeRF as a representative example. Similar trends are also observed across the other scenes. As shown in Table~\ref{tab:boc_analysis}, applying A-RDO to CAT-3DGS Pro with a triplane-based hyperprior (Row 1 → Row 2) has a minimal effect on the total file size, but improves the rendering quality (PSNR). Interestingly, A-RDO increases the size of the multi-scale planes ($\boldsymbol{\mathcal{P}}$) from 0.1M to 0.8M, but it does not significantly reduce the attribute size ($\boldsymbol{\mathcal{A}}$). This inspires us to take the opposite approach by reducing the number of channels in the triplane hyperprior from 72 to 8 (Row 2 → Row 3). This change results in negligible changes in both the rendering quality and total file size. However, it helps reduce the decoding and training complexity due to the lower channel count. Furthermore, replacing the triplane with our VM hyperprior (Row 3 → Row 4) producs a similar result, underscoring the advantage of adopting a simple hyperprior design. However, omitting the $\gamma(z')$ (denoted as M ($ch=8$)) produces a noticeable rate-distortion loss. % This adjustment preserves essential structural information while reducing parameter redundancy.} 
%Lastly, to evaluate the necessity of 3D spatial representation, we omit the vector from the third principal component,. As a result, the attribute size increases by approximately 6.4\%, underscoring the indispensability of the third principal component for a comprehensive hyperprior structure in 3D space.

%simplify the 
%This suggests that allocating more bits to $\boldsymbol{\mathcal{P}}$ does not yield proportional benefits for attribute compression. 

%We then 
%observe that the bitstream sizes for anchor attributes ($\boldsymbol{\mathcal{A}}$) and multi-scale planes ($\boldsymbol{\mathcal{P}}$) vary when A-RDO . However, it has a marginal impact on the total file size. However, it helps improve the rendering quality (PSNR) by 0.2dB. This is intuitively agreeable because it enables occasionally the rate term during the training process and emphasizes more the rendering quality.   
%Although A-RDO increases the size of the multi-scale planes ($\boldsymbol{\mathcal{P}}$) from 0.1M to 0.8M, it does not significantly reduce the attribute size ($\boldsymbol{\mathcal{A}}$). 

\subsubsection{Sampling Rate Optimization}
\label{exp:training_strategy}
\input{Sections/4-5_ablation_USRO_table}
Table~\ref{table:training_trick} presents the BD-rate savings of CAT-3DGS Pro under Uniform Sampling Rate Optimization (USRO). The results show that USRO consistently improves the rate-distortion performance across all the datasets. On the BungeeNeRF dataset, it achieves a 45.6\% BD-rate reduction, due to its training views being concentrated in a small central region. Recall that with USRO, less-viewed anchor points are also taken into account during the training process.

%% file: Sections/4-2_training_time_table.tex
\begin{table}[t]
    \begin{minipage}[t]{0.49\textwidth}
        % 上方表格 (TABLE III)
    \centering

    \caption{Training time comparison on BungeeNeRF.}
    % \vspace{2mm}
    \resizebox{\linewidth}{!}{
    \renewcommand{\arraystretch}{1.16} % Adjust the value as needed

    \begin{tabular}{c|c|c|c|c}
    \toprule
    \textbf{Method}   & \textbf{CAT-3DGS} & \textbf{CAT-3DGS Pro} &\textbf{HAC} &\textbf{ContextGS}\\
    \hline
      \textbf{Time (min)}& 224.0 & 73.5 & 45.3 & 104.2  \\
    
    \bottomrule
    
    \end{tabular}
    
    }
    \label{table:training_time}
    \end{minipage}
    \hfill
    \vspace{3.5mm}
\end{table}

%% file: Sections/4-3_decoding.tex
\begin{table}[t]
    \centering
    \renewcommand{\arraystretch}{1.2}

    \begin{minipage}[t]{0.499\textwidth}
        \centering
        \caption{Comparison of decoding time and rendering throughput.}
        \resizebox{\linewidth}{!}{
        \begin{tabular}{c|c|c|c|c|c|c}
        \toprule
        \multirow{2}{*}{\textbf{Method}} & \multirow{2}{*}{\textbf{Scene}} &\textbf{Anchor}& \multicolumn{3}{c|}{\textbf{Dec. Time (sec) $\downarrow$}} &  \textbf{Rendering}\\
        \cline{4-6}
        & &\textbf{Count} (K)&  $\boldsymbol{\mathcal{P}}$ & $\boldsymbol{\mathcal{A}}$ & \textbf{Total}& \textbf{Speed} (FPS) $\uparrow$\\
        \hline
        \multirow{2}{*}{CAT-3DGS}& \textit{room} &36.5& 11.9 & 2.3 & 14.2 & 127.0 \\
        & \textit{Amsterdam}&533.0&47.4 & 17.0 & 64.4 & 83.4 \\
        \hline
        \multirow{2}{*}{CAT-3DGS Pro}       
        &\textit{room} & 30.8 &1.0 & 1.1 & 2.1 & 132.7\\
        & \textit{Amsterdam}& 421.6 & 2.2  & 10.9 & 13.1 & 110.0 \\
        \hline
        \multirow{2}{*}{HAC} &\textit{room} &228.8& - & - & 6.6 & 103.1\\
        & \textit{Amsterdam}&483.5& - & - & 11.3 & 77.9  \\
        \hline
        \multirow{2}{*}{ContextGS} &\textit{room} & 119.4 & - & - & 2.6 & 92.2 \\
        & \textit{Amsterdam}& 360.3 & - & - &  8.4 &  113.1 \\
        \bottomrule
        \end{tabular}
        }
    \label{table:decode_time}
    \end{minipage}
    \hfill

\end{table}

%% file: Sections/4-4_ablation_bitstream_table.tex
\begin{table}[t]
        \centering
        \renewcommand{\arraystretch}{1.2}\
        \begin{minipage}[t]{0.45\textwidth}
            \caption{Analysis of A-RDO and the VM hyperprior on the \textit{Amsterdam} scene.}
            
            \resizebox{\textwidth}{!}{
            \begin{tabular}{c|c|c|c|c|c}
            \toprule
            \multirow{2}{*}{\textbf{Hyperprior}} & \multirow{2}{*}{\textbf{A-RDO}} & \textbf{PSNR} & \multicolumn{3}{c}{\textbf{Size (MB)}} \\ 
            \cline{4-6}
            & & (dB) & $\boldsymbol{\mathcal{P}}$ & $\boldsymbol{\mathcal{A}}$& Total \\ 
            \hline
            Triplane (ch=72) &  & 27.3 &0.1&19.2 & 23.8 \\ 
            \hline
            Triplane (ch=72) & \checkmark  &27.5 &0.8&18.4 &23.7 \\ 
            \hline
            Triplane (ch=8)  & \checkmark & 27.6 & 0.5&18.8  & 23.6 \\ 
            
            \hline
            Ours, VM (ch=8) & \checkmark  & 27.6&0.3&18.8 &23.5 \\
            \hline
            M (ch=8) & \checkmark  & 27.4 & 0.2& 20.0  & 24.6 \\ 
            \bottomrule
            \end{tabular}
    
        }
        \label{tab:boc_analysis}
        \end{minipage}
        \hfill
\end{table}

%% file: Sections/4-5_ablation_USRO_table.tex
\begin{table}[t]
    \centering
    \renewcommand{\arraystretch}{1.6}
    \begin{minipage}[t]{0.499\textwidth}
        \caption{BD-rate savings of USRO, with ViSRO as anchor.}
        \resizebox{\linewidth}{!}{
            \begin{tabular}{l|c c c c}
                \toprule
                \textbf{Scheme} & \textbf{Mip-NeRF 360} & \textbf{Tanks \& Temples} & \textbf{Deep Blending} & \textbf{BungeeNeRF} \\
                \hline
                USRO & -15.9\% & -9.8\% & -2.5\% & -45.6\% \\
                \bottomrule
            \end{tabular}
        }
        \label{table:training_trick}
    \end{minipage}
    \hfill
\end{table}

%% file: Sections/5_Conclusion.tex
\section{Conclusion}
CAT-3DGS Pro extends CAT-3DGS with an enhanced and accelerated rate-distortion-optimized 3DGS compression framework, achieving state-of-the-art coding performance with faster training and decoding. Through the proposed PCA-guided vector-matrix hyperprior, which simplifies the triplane structure in CAT-3DGS, we significantly reduce parameter redundancy and coding complexity while preserving a similar ability to model the distribution of anchor attributes. Additionally, the alternative optimization strategy (A-RDO) enables a more balanced rate-distortion trade-off and accelerates training. Lastly, our approach further enhances compression performance by refining the sampling rate optimization.
 
For future work, since current methods primarily focus on the entropy coding of anchor attributes while position information is simply stored in a 16-bit floating-point format, developing a unified compression framework for both position and attribute data in the ScaffoldGS representation is a promising approach to further improving rate-distortion performance.
% Additionally, as this framework is designed for per-scene optimization, developing a more generalizable framework applicable to new instances would enable more efficient and accelerated training.

% \textcolor{red}{For future work, we aim to explore a more effective joint compression strategy for both geometry and attribute information in 3D Gaussian Splatting. While current methods focus on optimizing appearance attributes, the redundancy in geometric structures, such as anchor positions, remains an open challenge. Developing a unified compression framework that efficiently encodes both geometry and attributes could further enhance rate-distortion performance.}